\pdfoutput=1

\documentclass[11pt]{article}
\usepackage{arabtex}
\usepackage{pdfpages}
\usepackage{subcaption}
\usepackage{utf8}
\setcode{utf8}
\def\endabstract{\egroup}

\usepackage[final]{acl}
\usepackage{times}
\usepackage{latexsym}
\usepackage{xcolor}
\definecolor{royalblue}{rgb}{0.25, 0.41, 0.88}
\definecolor{royalgreen}{HTML}{00B050}
\usepackage{amsmath}
\usepackage{amssymb}
\usepackage[most]{tcolorbox}
\usepackage{spverbatim}
\usepackage{array}
\usepackage{tablefootnote}
\newcommand{\icon}[1]{\includegraphics[height=2cm]{#1}}
\newcolumntype{P}[1]{>{\centering\arraybackslash}p{#1}}
\usepackage[T1]{fontenc}

\usepackage[utf8]{inputenc}

\usepackage{microtype}

\usepackage{graphicx}
\graphicspath{ {./imgs/} }

\usepackage{booktabs} 
\usepackage{array}
\usepackage{colortbl}
\usepackage{xcolor}
\usepackage{makecell}
\usepackage{graphicx}
\usepackage{multirow}
\usepackage{enumitem}
\usepackage{soul}

\newcommand{\Affilfont}{\fontsize{10}{11}\selectfont}

\title{
\protect \icon{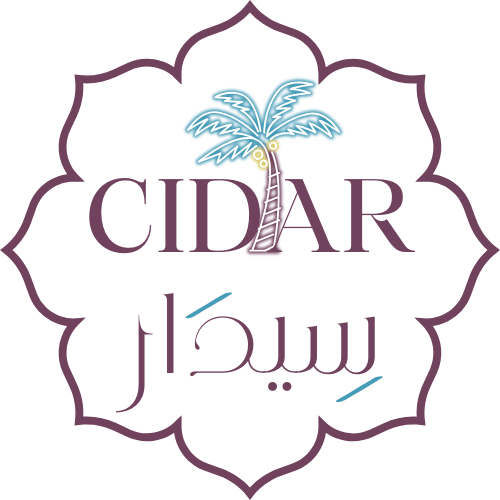}

\textsc{Cidar}: Culturally Relevant Instruction Dataset For Arabic
}

\author{
Zaid Alyafeai $^{1,*}$~~~ 
Khalid Almubarak  $^{2,*}$~~~ 
Ahmed Ashraf $^{3,*}$~~~
Deema Alnuhait $^{4,*}$~~~
\\
\textbf{Saied Alshahrani $^{5, 6}$~~~
Gubran A. Q. Abdulrahman $^1$~~~ 
Gamil Ahmed  $^{1,7}$~~~ 
}
\\
\textbf{
Qais Gawah $^{1}$~~~
Zead Saleh $^1$~~~
Mustafa Ghaleb $^{1,8}$~~~
Yousef Ali $^1$~~~
Maged S. Al-Shaibani $^1$~~~
}
\\
\Affilfont{$^1$ King Fahd University of Petroleum and Minerals (KFUPM) 
$^2$ Prince Sattam bin Abdulaziz University (PSAU)}\protect\\
\Affilfont{$^3$ ARBML
$^4$ University of Illinois Urbana-Champaign
$^5$ Clarkson University $^6$ University of Bisha} 
\protect\\
\Affilfont{
$^7$ Interdisciplinary Research Center for Smart Mobility and Logistics (IRC-SML), KFUPM}
\protect\\
\Affilfont{
$^8$ Interdisciplinary Research Center for Intelligent Secure Systems (IRC-ISS), KFUPM
}
}

\begin{document}
\maketitle
\def\thefootnote{*}\footnotetext{Equal contribution. Corresponding author: Zaid Alyafeai, email: \nolinkurl{g201080740@kfupm.edu.sa}}
\def\thefootnote{\arabic{footnote}}
\begin{abstract}
Instruction tuning has emerged as a prominent methodology for teaching Large Language Models (LLMs) to follow instructions. However, current instruction datasets predominantly cater to English or are derived from English-dominated LLMs, resulting in inherent biases toward Western culture. This bias significantly impacts the linguistic structures of non-English languages such as Arabic, which has a distinct grammar reflective of the diverse cultures across the Arab region. This paper addresses this limitation by introducing \textsc{Cidar}\footnote{\textsc{Cidar}: \href{https://hf.co/datasets/arbml/CIDAR}{https://hf.co/datasets/arbml/CIDAR}.} \textit{the first open Arabic instruction-tuning dataset culturally-aligned by human reviewers}. \textsc{Cidar} contains 10,000 instruction and output pairs that represent the Arab region. We discuss the cultural relevance of \textsc{Cidar} via the analysis and comparison to other models fine-tuned on other datasets. Our experiments show that \textsc{Cidar} can help enrich research efforts in aligning LLMs with the Arabic culture. All the code is available at \href{https://github.com/ARBML/CIDAR}{https://github.com/ARBML/CIDAR}.
\end{abstract}

\section{Introduction}

The need for Natural Language Processing (NLP) applications has exploded in an era of unprecedented linguistic interaction between humans and machines. As these applications strive for greater inclusivity and effectiveness across diverse linguistic landscapes, the need for datasets that reflect the cultural differences and linguistic peculiarities of specific regions becomes increasingly important. In the context of Arabic language understanding, the challenge lies not only in linguistic complexity but also in capturing the rich cultural fabric that shapes communication in the Arab world.

\begin{figure}[!htp]
    \centering
    \includegraphics[scale = 0.355]{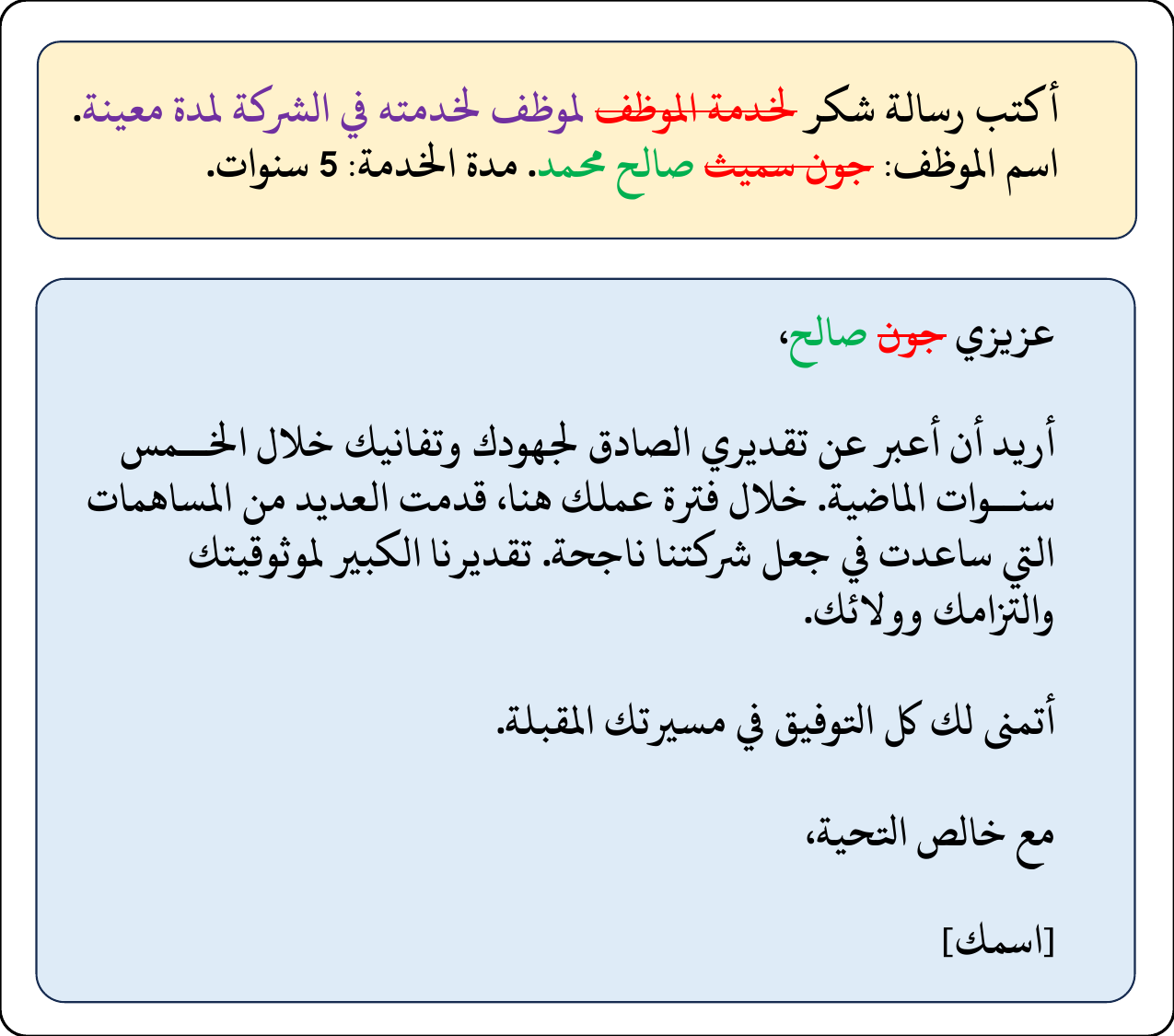}
    \caption{An example of our localization procedure in \textsc{Cidar} of a given (\colorbox[HTML]{FFF2CC}{instruction}, \colorbox[HTML]{DEEBF7}{output}) pair. We show, in colors, the \textcolor[HTML]{7030A0}{grammatical} and \textcolor{royalgreen}{cultural} modifications.}
    \label{fig:annot_example}
    \vspace{-9pt}
\end{figure}

In the past year, many language models have been pre-trained and instruct-tuned for Arabic, like \textsc{Jais} \cite{sengupta2023jais}, and \textsc{AceGPT} \cite{huang2023acegpt}. All these models have been trained on a large corpus of Arabic text and then fine-tuned to respond to users' instructions. However, such efforts do not release high-quality instruction datasets to be openly used for research. Moreover, they use a lot of machine-translated or machine-generated instruction datasets without further human review or audit, disregarding the consequences of using such poor, distorted, and misaligned instructions.

In this paper, we introduce \textsc{Cidar}, the \emph{first} open instruction-tuning dataset that has gone through extensive review and localization (see Figure \ref{fig:annot_example}) crafted for instructional tuning in Arabic. In the subsequent sections, we delve into the dataset creation process, elucidating the methodology employed to navigate the delicate balance between linguistic accuracy and cultural relevance. The paper also discusses the potential applications of \textsc{Cidar} in enhancing the performance of LLMs, shedding light on its role in bridging the gap between language understanding and cultural context within the realm of Arabic instruction-tuning. Mainly, we compare fine-tuning on a translated dataset and a localized dataset, i.e., \textsc{Cidar}. Ultimately, \textsc{Cidar} dataset stands as a testament to the evolving landscape of NLP research, advocating for the integration of cultural context as an essential component in the development of LLMs tailored for specific linguistic communities, like the Arab world.





\section{Related Work}
\label{sec:2}

Many efforts have been made to create numerous instruction datasets, especially for English; some are generated by LLMs like Stanford Alpaca \citep{alpaca}, Databricks’ Dolly \citep{DatabricksBlog2023DollyV2}, and \textsc{Self-Instruct} \citep{wang2023selfinstruct}, whereas others are human-generated with templates like Flan collections \citep{wei2022finetuned-flan, longpre2023-flan}, P3 \citep{bach2022promptsource-P3}, and \textsc{Natural Instructions} \citep{naturalinstructions}.  

In the following subsections, we briefly discuss the Arabic instruction-tuning datasets and their data collection approaches in a multilingual and monolingual context.


\subsection{Multilingual Instruction-tuning Datasets}
\citet{muennighoff2022crosslingual} 
presented xP3 (Crosslingual Public Pool of Prompts) as an extension of the P3 dataset \citep{sanh2022multitask-p3}, where the authors applied English prompts across 16 NLP tasks for 46 languages, including Arabic. Later, the authors introduced a much larger version called xP3x (Crosslingual Public Pool of Prompts eXtended), in which they extended the English prompts to 277 languages, including Arabic and ten of its dialects. Despite their large sizes, these datasets exhibit limited variation due to their reliance on prompt template structure and their emphasis on classical NLP tasks such as translation, question answering, text classification, text summarization, and other tasks. 




\citet{Chen_MultilingualSIFT_Multilingual_Supervised_2023} constructed \textsc{MultilingualSIFT} (Multilingual Supervised Instruction Fine-tuning) datasets, by translating instructions for 11 languages, including Arabic. The authors translated these three training datasets: Alpaca-GPT4 \citep{peng2023instruction}, Evol-Instruct \citep{xu2023wizardlm}, and ShareGPT \citep{zheng2023judging}, from English to Arabic using GPT-3.5 Turbo. For Alpaca-GPT4, they directly translated the instructions and responses, while for Evol-Instruct and ShareGPT, they translated the instructions and used them to generate the responses. Furthermore, the authors translated two evaluation datasets, \citep{hendrycks2021measuring} and Vicuna-80 \citep{zheng2023judging}, using the same above-mentioned approach.\vspace{2pt}

\citet{wang2022supernaturalinstructions} introduced \textsc{Super-NaturalInstructions} (\textsc{Sup-NatInst}) as the first benchmark of 1,616 diverse NLP tasks, along with their expert-written instructions. It covers nearly 76 distinct task types like text classification, extraction, rewriting, and composition, spanning 55 languages. It includes 80.3K Arabic instructions for 16 Arabic NLP tasks like text translation and sentence perturbation generation, yet the number of Arabic NLP tasks is underrepresented compared to other languages like Spanish (43 tasks), Japanese (40 tasks), and Persian (34 tasks). \vspace{2pt}

\citet{li2023bactrianx} presented Bactrian-X, a 3.4M instruction-response pair for 52 human languages, including Arabic, with around 65.4K pairs. The authors \textit{only} translated selected instructions from Alpaca \cite{alpaca} and Dolly \citep{DatabricksBlog2023DollyV2}, using Google Translate\footnote{Google Translate: \href{https://translate.google.com}{https://translate.google.com}.} to Arabic. After that, they generated responses for these selected translated instructions using GPT-3.5 Turbo. 



\citet{upadhayay2023taco} introduced the Multilingual Instruction-Tuning Dataset (MITD), which is composed of the translation of Alpaca-GPT4 \citep{peng2023instruction}, Dolly \citep{DatabricksBlog2023DollyV2}, and Vicuna Benchmark \citep{vicuna2023} in 132 languages, including Arabic, using Google Cloud Translation\footnote{Google Cloud Translation: \href{https://cloud.google.com/translate}{https://cloud.google.com}.}. Despite the authors' acknowledgment that their translations are prone to \emph{translationese}, where the translated texts deviate from the native language norms due to many factors like the overly literal translation and unusual phrases or word choices, they \emph{only} performed a manual evaluation of the translation quality for four language, unsurprisingly Arabic was not one of them.

\citet{köpf2023-openassistant} released OpenAssistant Conversations (OASST1), a human-generated and human-annotated assistant-style conversation dataset consisting of 161.4K messages in 35 human languages, including Arabic, resulting in over 10K complete and fully annotated conversation trees. This was a product of a worldwide crowd-sourcing effort involving over 13.5K volunteers. The Arabic portion has only 666 data samples. 

\subsection{Arabic Instruction-tuning Datasets}

In the context of training Arabic-specific LLMs, a few attempts were made to create Arabic instruction-tuning datasets. However, most of these datasets are closed (not publicly released).

\citet{chen2023phoenix} released their instruct-tuned model \textsc{Phoenix} using three groups of instructions: collected multi-lingual instructions, post-translated multi-lingual instructions, and self-generated user-centered multi-lingual instructions. Specifically, in the post-translated multi-lingual instructions, the authors translated Alpace instruction and output pairs \citep{alpaca} using GPT-4 to Arabic, and sometimes they generated responses for the GPT-4 translated instructions using GPT-3.5 Turbo for alleviating the unavoidable translation issues.  


\citet{naseej} instruct-tuned their model \textsc{Noon} using a collection of Arabic instructions from different datasets, such as Alpaca-GPT4 \cite{peng2023instruction}, Databricks’ Dolly \cite{DatabricksBlog2023DollyV2}, TruthfulQA dataset \cite{lin2021truthfulqa}, Grade School Math dataset \cite{cobbe2021gsm8k}, and Arabic arithmetic problems generated using GPT-3.5 Turbo. 

\citet{sengupta2023jais} also instruct-tuned their model \textsc{Jais} using a translated collection of instructions to Arabic from various instructions-tuning datasets, such as \textsc{Super-NaturalInstructions}, \cite{wang2022supernaturalinstructions} 
Unnatural \cite{honovich-etal-2023-unnatural}, NaturalQuestions \cite{nq}, Alpaca \cite{alpaca}, HC3 \cite{guo2023close}, Databricks’ Dolly \cite{DatabricksBlog2023DollyV2},
Basic-Conv\footnote{ChatterBot Corpus: \href{https://chatterbot-corpus.readthedocs.io}{https://chatterbot-corpus.docs.io}
}, Bactrian-X \cite{li2023bactrianx} and enriched the collection with Arabic examples from xP3 \cite{muennighoff2022crosslingual}. The authors also formatted the AraNER \cite{ANERsys} to the instruction-response format. Furthermore, the authors created two unreleased datasets with instruction-response pairs for the United Arab Emirates (UAE) and the region: NativeQA-Ar and SafetyQA-Ar.



\citet{huang2023acegpt} as well instruct-tuned their model \textsc{AceGPT} using instructions compiled from some open-source datasets, like Alpaca \cite{alpaca}, Alpaca-GPT4 \cite{peng2023instruction}, Evol-Instruct \cite{xu2023wizardlm}, Code-Alpaca \cite{codealpaca}, and ShareGPT \cite{zheng2023judging}, and translated the questions from English to Arabic and re-generated the responses using GPT-4. Moreover, the authors fine-tuned their model using native Arabic instructions collected from the question-answering platform Quora\footnote{Quora: \href{https://www.quora.com}{https://www.quora.com}} as localized instructions and generated responses for these instructions using GPT-4. Plus, the authors introduced a translated version of Arabic-Vicuna-80\footnote{Arabic-Vicuna-80: \href{https://hf.co/datasets/FreedomIntelligence/Arabic-Vicuna-80}{https://hf.co/datasets/FreedomIntelli-gence/Arabic-Vicuna-80}.} to conduct a human evaluation study. 


\citet{almazrouei-etal-2023-alghafa} lately instruct-tuned a few models using multiple machine-translated Arabic instruction-tuning datasets, including xP3 \cite{muennighoff2022crosslingual}, Bactrian-X \cite{li2023bactrianx}, Alpaca \cite{alpaca}, and UltraChat \cite{ding2023enhancing}. The authors also performed human evaluations of the fine-tuned models on multiple categories, including education, health, technology, history, creativity, oil, and gas.

\citet{Alpaca_arabic_instruct} released the \textit{only} open-source monolingual, Arabic instruction-tuning dataset, which is poorly translated from Alpaca dataset \cite{alpaca} to the Arabic language using Google Translate without cultural alignment or even a simple translation error checking.







\begin{figure*}[!htp]
  \centering
\includegraphics[scale=0.19]{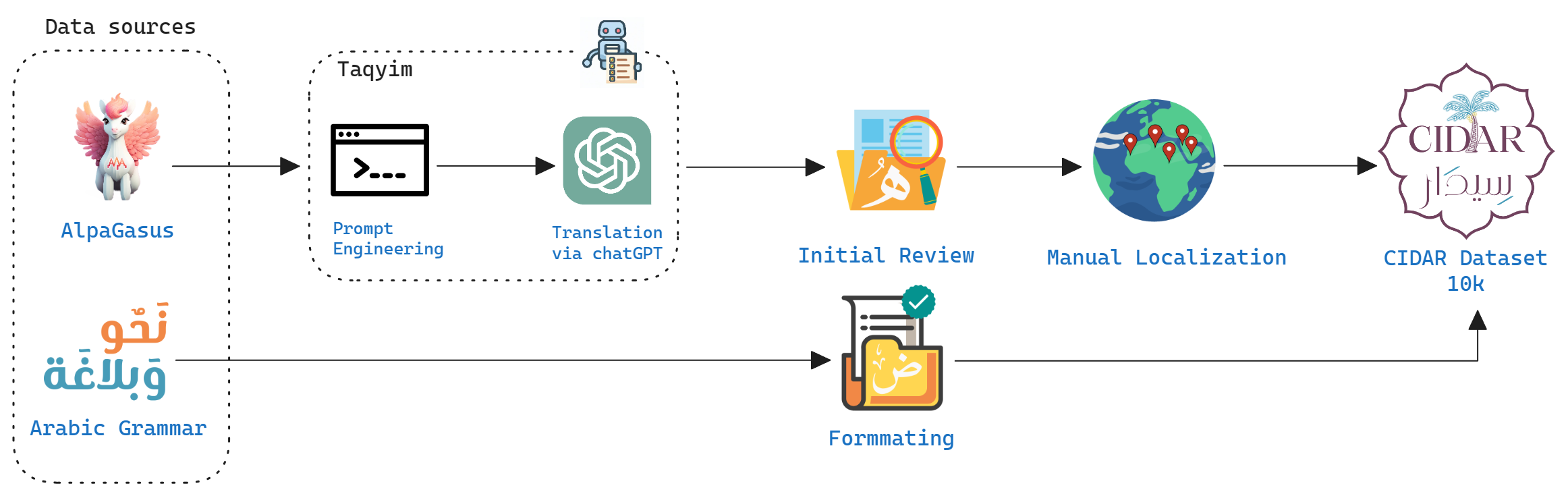}
 \caption{Workflow diagram of \textsc{Cidar}’s data collection pipeline, illustrating each pipeline phase and its components.}
    \label{fig:dc}
\end{figure*}
\section{Issues of Arabic Instruction Datasets}
\label{sec:3}
Two main approaches were addressed in the previous literature for creating Arabic instruction-tuning datasets: the full translation of both instruction-response pairs using Machine Translation (MTs) tools and the translation of instructions, then generating responses using LLMs like GPT-4. However, each creation or generation approach of the Arabic instruction-tuning datasets has serious drawbacks that we discuss next.

\subsection{MTs-\textit{related} Issues}
One harmful drawback of the current instruction-tuning datasets' creation approaches is the poor, naive, and direct translation of  English instruction-response pairs to Arabic without human intervention or supervision using off-the-shelf MT tools like Google Translate, which is widely known for their social problems like gender, cultural, and religious biases and stereotypes \citep{Marcelo-2022-Genderbias, ullmann_saunders_2021-Genderbias, Lopez-Medel2021-Genderbias, Chen-et-al-2021-Genderbias, naik2023reducing, alshahrani-etal-2022-roadblocks}. Many researchers have repeatedly stressed how such unguided translations are not only prone to various linguistic and grammatical errors, detrimental outcomes, cultural misalignment (favoring the Western culture), and representational harm to native speakers (unrepresentative content) but also introduce negative performance implications of models trained on them 
\citep{stanovsky-etal-2019-evaluating, habash-etal-2019-automatic, DasAlok, Agrawal-2023-Translation-Tools, alshahrani-etal-2023-performance}. 





\subsection{LLMs-\textit{related} Issues}
The other hazardous drawback of the current instruction-tuning datasets' creation approaches is the unvetted, unchecked, and unsupervised translation of instruction-response pairs from English to Arabic or the generation of responses for the previously translated instructions, all using LLMs like GPT-3.5 Turbo or GPT-4 without paying attention to the consequences. Many research studies have underscored various risks, threats, and controversies in LLMs, for example, research studies like \cite{paullada2021data, Wach_2023_chatGPT, thakur2023unveiling, naous2023having, dong2023probing, human-like-content-llms} accentuated that most commonly used LLMs could exhibit a wide spectrum of biases, privacy, and security hazards, ethical questions, hallucination, and could create a damaging or deceptive content of certain group. Besides, LLMs could generate content (e.g., responses) that suffer cultural misalignment and cultural incongruencies, leading to culturally unaligned, undiverse, untruthful, and unrepresentative outputs \cite{prabhakaran2022cultural, alshahrani-etal-2022-learning, kasirzadeh2022conversation, cetinic2022myth, bang2023multitask, yu2023large, masoud2023cultural, ji2024ai}.

\begin{table*}[!htp]
    \caption{Comparison between translated \textsc{AlpaGasus} and \textsc{Cidar} regarding names and countries using Word Clouds. In \textsc{AlpaGasus}, the top locations are the United States (\<الولايات>) and New York (\<نيويورك>), \vspace{-4.4pt}and the top names are John (\<جون>) and Marry (\<ماري>)\vspace{-6pt}, while in \textsc{Cidar}, after our localization, the top locations are Yemen (\<اليمن>) and Egypt (\<مصر>), and the top names are Muhammad (\<محمد>) and Sarah (\<سارة>). 
    }
    \label{tab:loc-pers}
    \begin{center}
    \begin{tabular}{c|c}
        \textbf{Locations in \textsc{AlpaGasus}} & \textbf{Locations in \textsc{Cidar}} \\
        \includegraphics[width=0.4\textwidth]{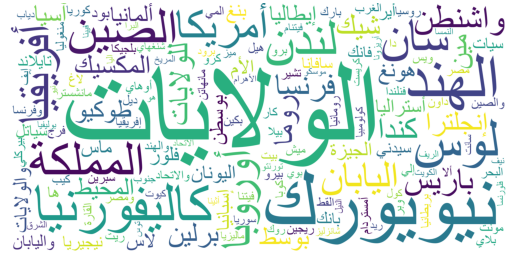} & \includegraphics[width=0.4\textwidth]{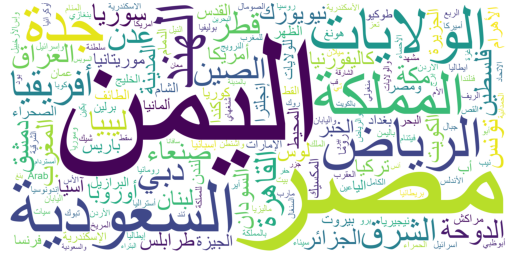}  \\ \hline
        \textbf{Names in \textsc{AlpaGasus}} & \textbf{Names in \textsc{Cidar}} \\
         \includegraphics[width=0.4\textwidth]{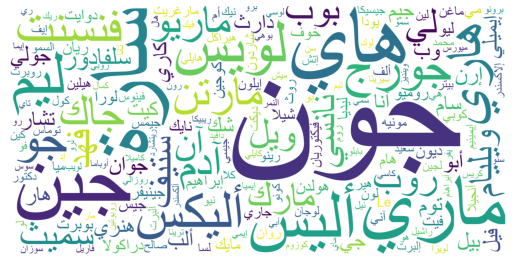}& \includegraphics[width=0.4\textwidth]{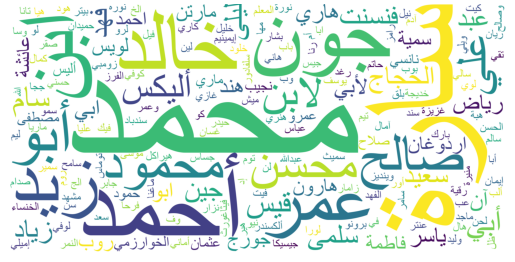} 
    \end{tabular}
    \end{center}
\end{table*}

\section{\textsc{Cidar}}
In this paper, we introduce \textsc{Cidar} which was constructed from two different sources. First, we use the \textsc{AlpaGasus} dataset \footnote{\textsc{AlpaGasus}: \href{https://hf.co/datasets/mlabonne/alpagasus}{https://hf.co/mlabonne/alpagasus}} reproduced from the work by \cite{chen2023alpagasus} which is a high-quality dataset filtered from the Stanford Alpaca dataset \cite{alpaca}. \textsc{AlpaGasus} contains more than 9K instruction, input, and output triplets. We translate 9,109 of the data to Arabic using ChatGPT (GPT-3.5 Turbo). Then,  we append it with around 891 questions and answers about the Arabic language and Grammar crawled from AskTheTeacher website\footnote{AskTheTeacher: \href{https://learning.aljazeera.net/ar/asktheteacher}{https://aljazeera.net/ar/asktheteacher}}. Figure \ref{fig:dc} highlights the main procedure for our data collection process. Next, we explain our approach to construct \textsc{Cidar} further.

\subsection{Machine Translation}
To translate \textsc{AlpaGasus}, we tested with different prompts to observe which one gave the best results.  Also, We use the Taqyim library \cite{alyafeai2023taqyim} to translate all the examples in \textsc{AlpaGasus} using GPT-3.5 Turbo. Initially,  We tested with direct translation of instruction, input, and output triplets but that did not give the best results. Hence, we concatenated the instructions and input. Another challenge encountered was ChatGPT translating coding blocks. Consequently, we had to explicitly instruct it to ignore coding blocks. We also append the instruction and output with \textit{User}, and \textit{Bot}, respectively, as in the following example:


\begin{tcolorbox}
\textit{You are given a conversation between a user and a bot, translate the full conversation into Arabic. Don't translate any coding blocks.}

\textbf{User}: 
Given the context, identify a suitable word to complete the sentence. The sun feels so <mask> today, I just want to sit here and relax.

\textbf{Bot}: warm.
\end{tcolorbox}

\subsection{Initial Review}
After translating our seed dataset, we noticed some initial problems. Therefore, we followed multiple steps to fix these machine translation issues:

\begin{itemize}
    \item Fix instructions or outputs that contain a large number of the English alphabet.
    \item Fix empty fields of instructions or outputs.
    \item Fix manually instructions that had wrong first words that are not in the correct form of an instruction.
\end{itemize}

The main goal of this step is to observe the current problems in the dataset to initialize the guidelines for the annotators.

\subsection{Localization}
\label{localization}
After fixing the initial issues with translation, we prepare our dataset to be manually reviewed. To simplify the annotation process, we created a web-based Annotation Tool (see Appendix \ref{appx:Annotation-App}), where reviewers were instructed to fix two main issues: 

\begin{itemize}
    \item \textbf{Linguistic Issues}: Some words might not be translated correctly, especially at the beginning of each instruction; we want all the statements to start with an instruction. For example, we should replace \<خلاصة> \vspace{-5pt}(summary) with \<لخّص> (summarize). Also, some instructions might be specific to English. The annotators are asked to provide their corresponding examples in Arabic. 

    \item \textbf{Cultural Relevance}: Some examples in the original English Alpaca might contain examples that represent Western cultures. We want to replace them with instructions that represent the Arab region and its culture. For instance, the name\vspace{-2pt} \<جون سميث> (John Smith) should be replaced by an Arabic name like \<علي خالد> (Ali Khalid).
\end{itemize}

\begin{figure*}[!htp]
    \centering
    \includegraphics[width=\linewidth]{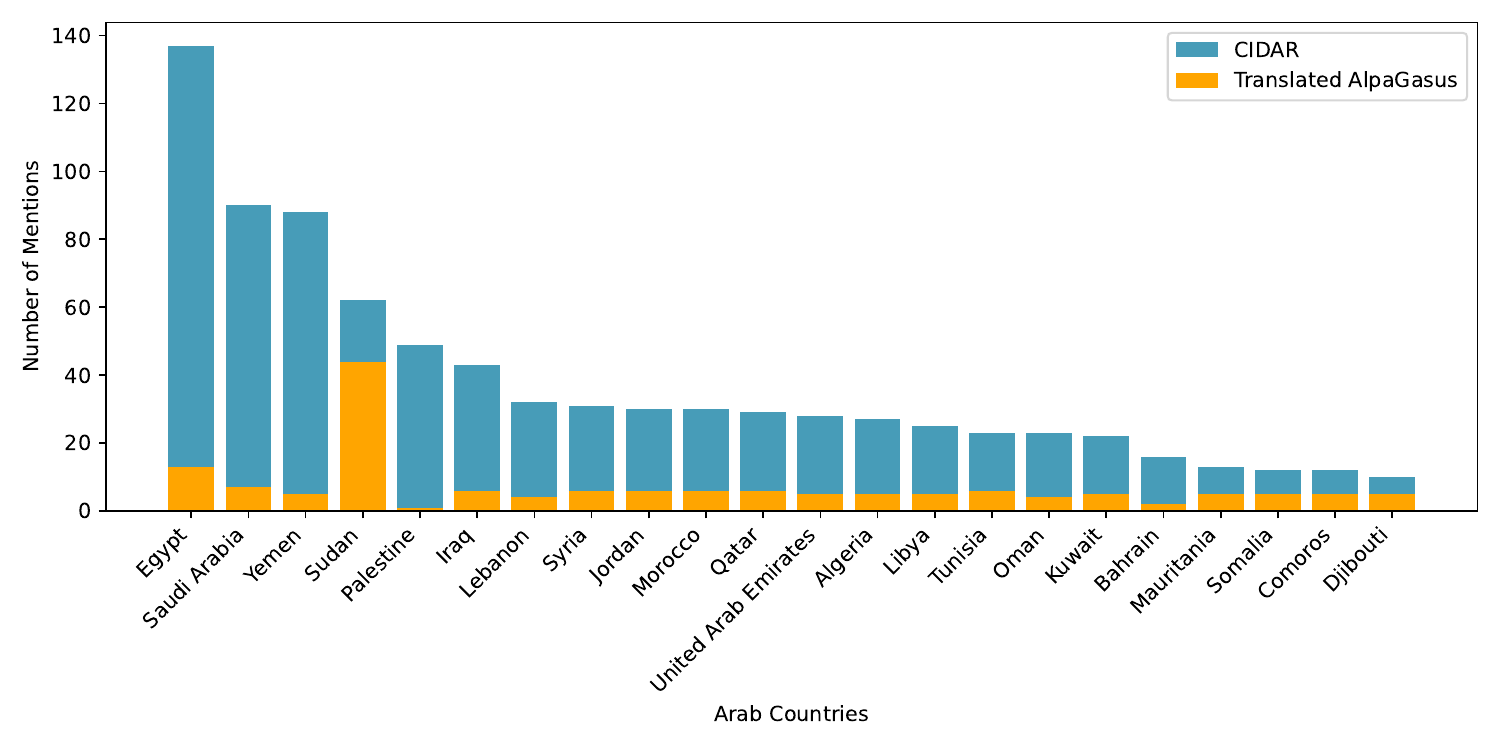}
    \caption{Number of mentions of every Arab country in both \textsc{Cidar} and translated \textsc{AlpaGasus} datasets.}
    \label{fig:country_mentions}
    \vspace{-3pt}
\end{figure*}
\section{Dataset Analysis}
In our data-gathering process, around 12 contributors participated in reviewing the dataset. Hence, in total, we have 10,000 instruction and output pairs that went under review. In this section, we compare between \textsc{Cidar} and the initial translated \textsc{AlpaGasus}. Through such analysis, we aim to emphasize the importance of manual revision and cultural alignment of machine-generated data.


\subsection{Modifications}
 Table \ref{tab:mod} shows the number of modifications in \textsc{Cidar} concerning the instructions, outputs, or either of them. From 9,109 instruction-response pairs in \textsc{AlpaGasus}, there were around \textit{64.5\%} of them that required a modification to be included in \textsc{Cidar}. These modifications are either due to a linguistic error or cultural irrelevance.
\begin{table}[!htp]
    \centering
    \begin{tabular}{l|c}
    \textbf{Modifications} & \textbf{\# Samples} \\ \hline \hline
        Instructions & 3,202 \\
        Outputs & 4,879\\
        Instructions or Outputs  & 5,871 \\ \hline
    \end{tabular}
    \caption{Number of modified instructions and outputs from the original \textsc{AlpaGasus} using manual review.}
    \label{tab:mod}
    \vspace{-3pt}
\end{table}

\subsection{Locations and Names}
The translated \textsc{AlpaGasus} dataset contains a lot of Western names and countries. To calculate how much \textsc{Cidar} mitigates that, we use Named Entity Recognition (NER) to extract the tokens that represent persons and locations. we use a fine-tuned CAMeLBERT \cite{inoue2021interplay} model on NER\footnote{CAMeLBERT NER: \href{https://hf.co/CAMeL-Lab/bert-base-arabic-camelbert-mix-ner}{https://hf.co/CAMeL-Lab/bert-base-arabic-camelbert-mix-ner}} to extract the names of persons and countries in both \textsc{Cidar} and the translated \textsc{AlpaGasus}. 
In Table \ref{tab:loc-pers}, we draw a comparison between locations and persons in both datasets using word cloud visualizations. 
We can see that the majority of locations and names in \textsc{Cidar} are from the Arab region.

\begin{figure*}
 \begin{subfigure}{0.45\textwidth}
    \centering
    \includegraphics[width=\textwidth]{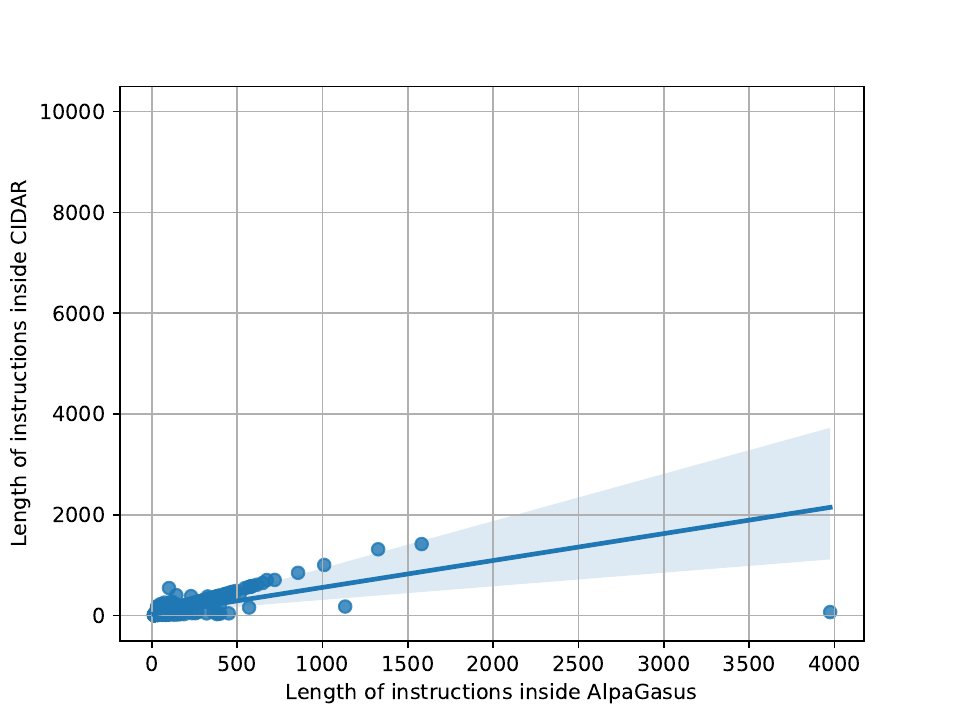}
    \caption{Comparison of instruction lengths.}
    \label{fig:subfig1}
  \end{subfigure}
  \hfill
  \begin{subfigure}{0.45\textwidth}
    \centering
    \includegraphics[width=\textwidth]{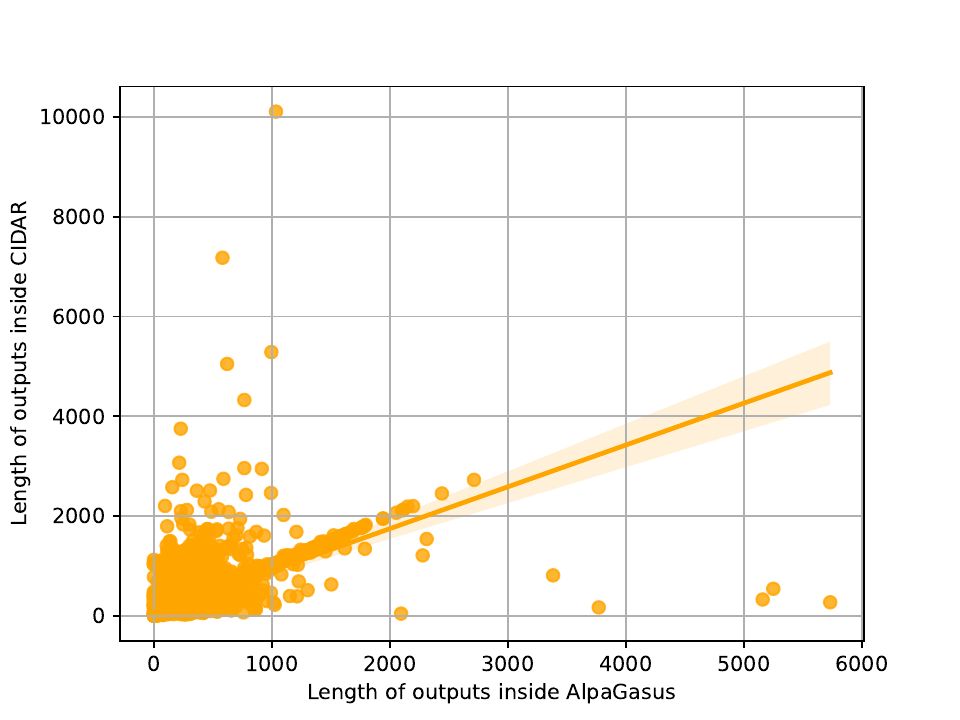}
    \caption{Comparison of output lengths.}
    \label{fig:subfig2}
  \end{subfigure}

  \caption{Comparison between \textsc{Cidar} and translated \textsc{AlpaGasus} in terms of instruction (Left) and output (Right) lengths. Noticeably, the length of outputs increased in \textsc{Cidar} due to the possible reviewers' rewriting of outputs.}
  \label{fig:lengths}
\end{figure*}
\subsection{Countries}
In Figure \ref{fig:country_mentions}, we highlight the distribution of (instruction, output) pairs that contain Arab countries. We observe a huge superiority for \textsc{Cidar} over the translated \textsc{AlpaGasus} in terms of mentioning Arab countries. In \textsc{Cidar}, the mentions of Arab countries have increased noticeably after our localization. While, in \textsc{AlpaGasus}, the mentions of Arab countries are mostly around ten mentions for most countries, except for Sudan (\<السودان>). This highlights the importance of \textsc{Cidar} in representing the region. Note that Sudan is considered an outlier because many food recipes contain peanuts as an ingredient, which is translated \vspace{-3pt}to \<فول سوداني> (Sudanese Bean) in Arabic.  

\subsection{General Topics}
\begin{figure}[!htp]
    \centering
    \includegraphics[width =0.38\textwidth]{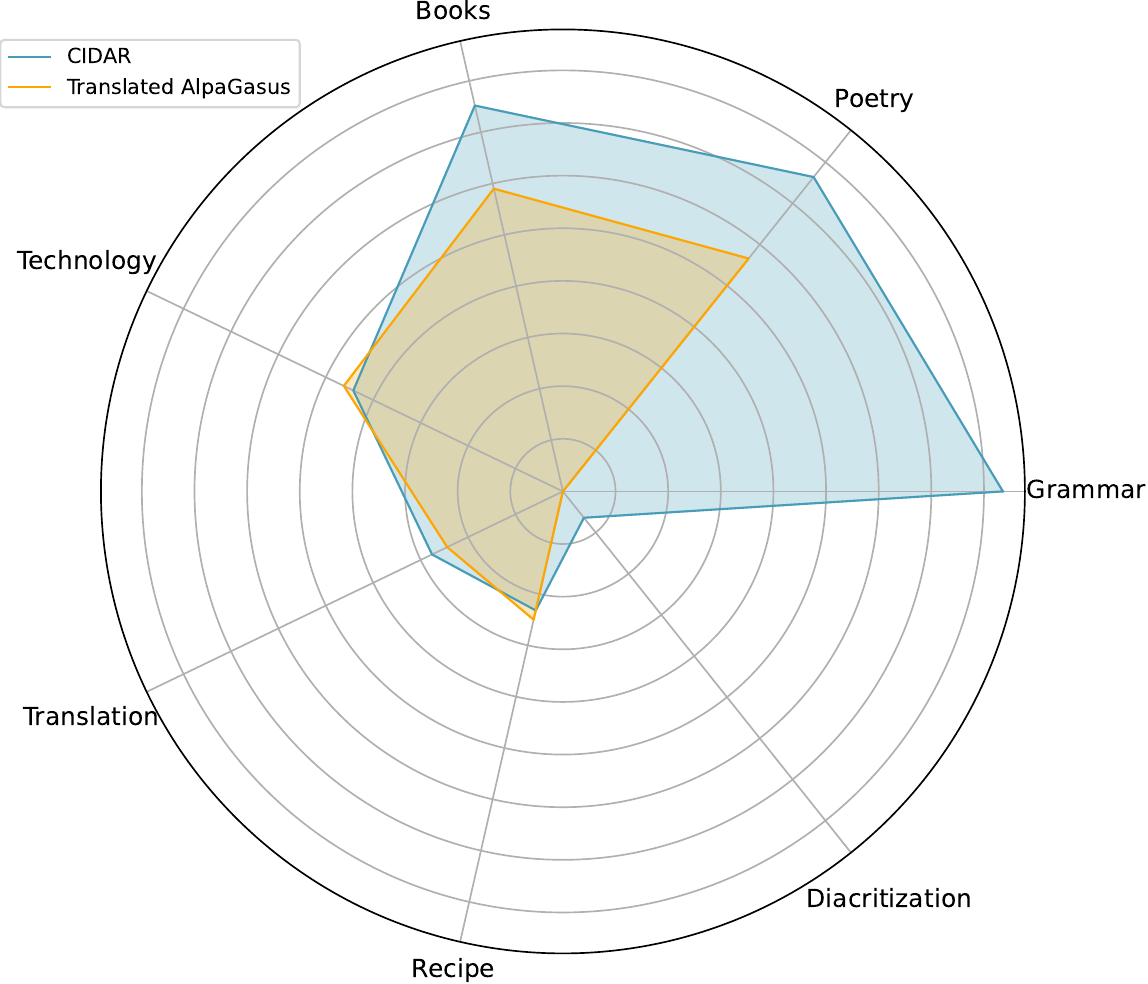}
    \caption{Comparison between \textsc{Cidar} and translated \textsc{AlpaGasus} datasets regarding the covered topics.}
    \label{fig:topics}
\end{figure}

We use keyword-based search to extract how many (instruction, output) pairs contain a specific topic. In Figure \ref{fig:topics}, we observe that, in general, \textsc{Cidar} covers a wider range of topics, including Arabic-specific tasks such as Arabic grammar and diacritization, which are largely missed in \textsc{AlpaGasus}.


\subsection{Annotation Lengths}
In Figure \ref{fig:lengths}, we compare the length of instructions and outputs between \textsc{Cidar} and translated \textsc{AlpaGasus} before and after our review. We highlight 
fewer changes in terms of instructions compared to outputs after the review. This is expected because sometimes the reviewer might re-write the whole output depending on changing a few words in the instruction. For example, if an instruction asks to find the best tourist places in a given US state, the reviewer will \textit{likely} change one word in the instruction and completely rewrite the whole output, which might result in a longer output.

\section{Evaluation}
We employed \textsc{AceGPT}-7B, a variant of LLaMA-7B fine-tuned on Arabic datasets \cite{huang2023acegpt}, as the base model. This model was further fine-tuned using two instruction datasets, \textsc{Cidar} and an Arabic-translated version of \textsc{AlpaGasus}, to assess their adaptability in culturally and regionally nuanced contexts.
This study compares the following three variants of \textsc{AceGPT} across diverse cultural and regional scenarios, models defined as: 
\begin{enumerate}

    \item \textbf{\textsc{AceGPT}\textbackslash \textsc{Cidar}}
    : A fine-tuned variant of the pre-trained \textsc{AceGPT}-7B model on our culturally aligned dataset, \textsc{Cidar}.
    
    \item \textbf{\textsc{AceGPT}\textbackslash \textsc{AlpaGasus}}
    : A fine-tuned variant of the pre-trained \textsc{AceGPT}-7B model on translated \textsc{AlpaGasus} dataset.  
    
    \item \textbf{\textsc{AceGPT}\textbackslash \textsc{Chat}}\footnote{\textsc{AceGPT}\textbackslash\textsc{Chat}: \href{https://huggingface.co/FreedomIntelligence/AceGPT-7B-chat}{https://huggingface.co/FreedomIntell-igence/AceGPT-7B-chat}.}: The instruct-tuned variant of \textsc{AceGPT}-7B model released by the original authors \cite{huang2023acegpt}.  
\end{enumerate}

We fine-tuned \textsc{AceGPT}\textbackslash \textsc{AlpaGasus} and \textsc{AceGPT}\textbackslash \textsc{Cidar} models, using supervised fine-tuning (SFT) with the Quantized Low-Rank Adaptation (QLoRA) quantization technique, as outlined in \cite{qlora}. Detailed specifications of the fine-tuning and inference hyper-parameters are provided in Appendix \ref{appx:params}. In Table \ref{tab:num_instructions}, we compare the number of instructions used to fine-tune each model. 




\begin{table}[!htp]
    \centering
    \begin{tabular}{l|c}
    \textbf{Model} & \textbf{\# Innstructions} \\ \hline \hline
        \textsc{AceGPT}\textbackslash \textsc{Cidar} & 10,000 \\
        \textsc{AceGPT}\textbackslash \textsc{AlpaGasus} & 9,230\\
        \textsc{AceGPT}\textbackslash \textsc{Chat}  & 363,155 \\ \hline
    \end{tabular}
    \caption{Number of instructions used for fine-tuning each model in our evaluation.}
    \label{tab:num_instructions}
\end{table}

\begin{figure*}[!htp]
  \centering
\includegraphics[scale=0.63]{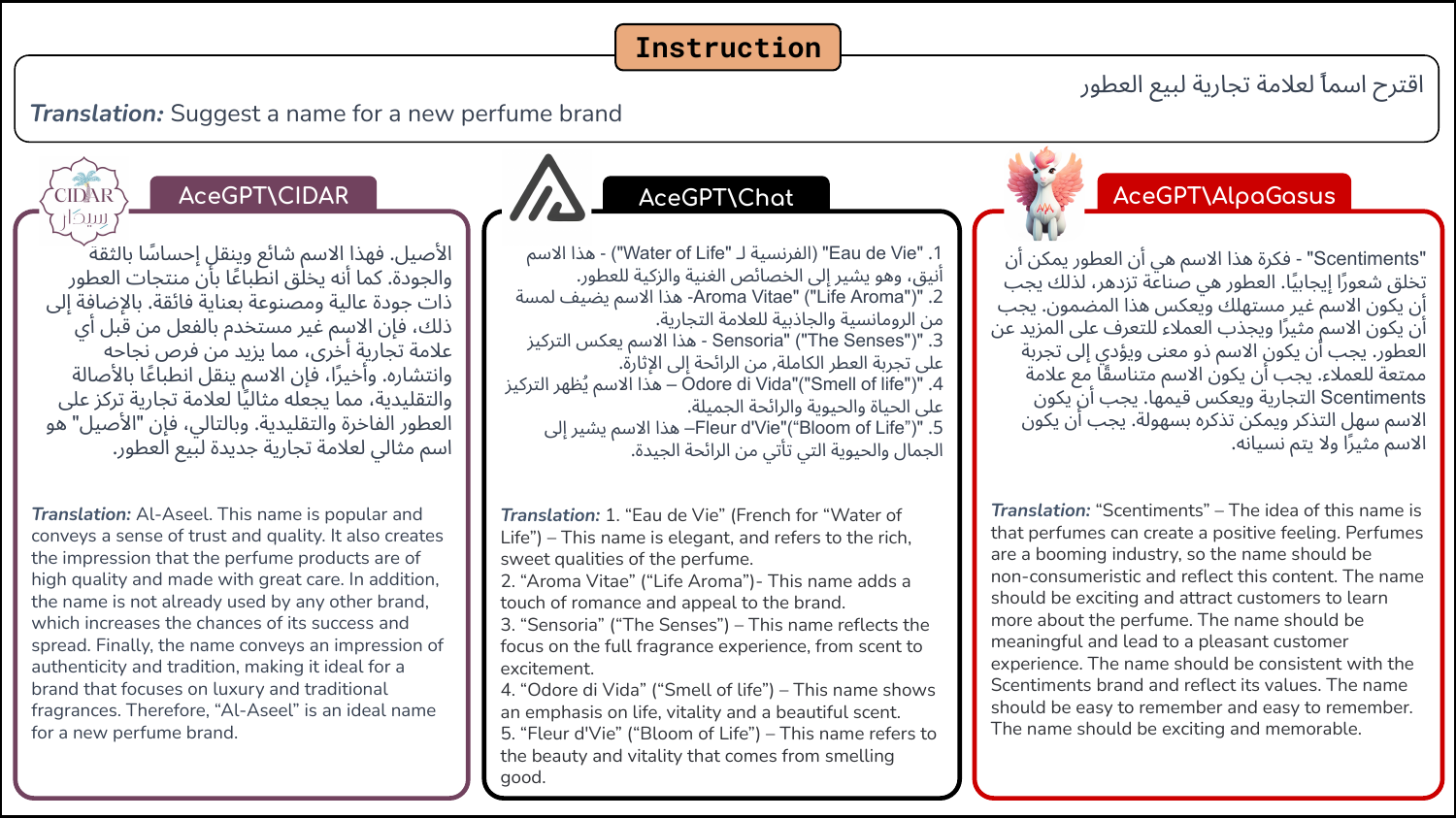}
\caption{Comparison between the outputs of the three evaluated models on a given instruction.} 
\label{fig:comparison}
\end{figure*}

 Figure \ref{fig:comparison} illustrates a qualitative example, showcasing the outputs of the three models on a given instruction. In this example, we want to know which model can capture the names that are related to the culture in the region. We observe that \textsc{AceGPT}\textbackslash \textsc{Cidar} demonstrated a marked improvement in aligning with Arabic culture by choosing a perfume name that is related to our region. In contrast, the \textsc{AceGPT}\textbackslash \textsc{AlpaGasus} model showed a tendency towards creating English names. We also observe that \textsc{AceGPT}\textbackslash \textsc{Chat} creates a list of suggestions of the names, even though this was not requested in the instruction. For more detailed examples, refer to Table \ref{table:finetune_param} in Appendix \ref{appx:outputs}.

\section{Social Impact and Limitations}
We aim to establish \textsc{Cidar} with the primary goal of incorporating rich Arabic content that authentically reflects our cultural values and the linguistic beauty of the language. Unlike much of the existing literature that relies on translated datasets or LLM-generated responses, which may encounter many challenges, as previously discussed, our focus is on preserving the integrity/quality of Arabic instruction. Moreover, the original Alpaca or \textsc{AlpaGasus} predominantly features Western cultural themes, such as food recipes, poems, tourist destinations, names, and countries. In our endeavor to curate \textsc{Cidar}, we have diligently ensured the inclusion of elements specific to our culture and traditions, encompassing Arabic linguistic nuances, narratives, tourism, names, culinary recipes, poetry, and countries. The open release of the dataset allows for fine-tuning LLMs that are cultural-aligned and can help with different domains. Our pilot study on fine-tuning \textsc{AceGPT} shows the huge impact such datasets can have in the region. 

That being said, \textsc{Cidar} still poses some limitations related to the data curation process. We summarize them as the following:
\begin{itemize}
    \item \textbf{Country Biases}: Localizing a given instruction usually depends on the nationality of the person annotating. 
    Often, annotators will prefer to add annotations related to the countries they were born in or currently residing in. 

    \item \textbf{Dataset Size}: The size of the dataset might limit its uses in large-scale instruction tuning. In our evaluation, we attempted to show that it helps to train on a culturally relevant dataset. 
    
    \item \textbf{Topics Covered}: In our data localization process, we tried to cover as many topics that are related to the culture of the region. We opted out of topics related to religion as it is considered a sensitive topic in the region. 

    \item \textbf{Dialects}: The Arabic language is not limited to Modern Standard Arabic (MSA). There are various Arabic dialects. Localization of data was limited to corrections of the translated text, which is mostly written in MSA, without incorporating multiple dialects.

    \item \textbf{Safety}: Due to the relatively small size of \textsc{Cidar}, the fine-tuned models on our dataset can show some degree of hallucinations, especially that it is not subjected to further alignment processes such as Reinforcement Learning from Human Feedback (RLHF) \cite{ouyang2022training}.
    
\end{itemize}
\section{Conclusion}
In conclusion, this paper introduces a significant contribution to the field of language model training by presenting \textsc{Cidar}, the \emph{first} open Arabic instruction-tuning dataset that is culturally-aligned by human reviewers. We highlight that the conventional approach of fine-tuning on machine-generated or machine-translated datasets has often resulted in biases favoring Western cultural nuances. Recognizing the unique grammar and cultural richness of the Arabic language, our dataset curation process aims to localize a given seed dataset, fostering a more authentic representation.

Through careful analysis and comparison with other models fine-tuned on other datasets, we demonstrate that \textsc{Cidar} serves as a pivotal resource for enriching research efforts in aligning Large Language Models (LLMs) with the Arabic culture. The experiments conducted not only validate the cultural relevance of our dataset but also highlight its potential to enhance the performance and understanding of LLMs within the Arabic linguistic and cultural context.

The availability of \textsc{Cidar} and the transparency of our dataset curation approach provide a foundation for future advancements in Arabic language model research. Researchers and developers can leverage this dataset to train models that better comprehend and respond to instructions within the cultural nuances of the Arab region. By sharing our code openly on GitHub, at \href{https://github.com/ARBML/CIDAR}{https://github.com/ARBML/CIDAR}, we encourage collaboration, further refinements, and broader contributions to the ongoing efforts to align language models with diverse cultural and linguistic contexts. Ultimately, \textsc{Cidar} stands as a valuable resource for advancing the inclusivity and effectiveness of language models in the Arabic-speaking world.  

\vspace{3pt}

\section*{Acknowledgements}
We would like to thank Maqsam\footnote{Maqsam: \href{https://maqsam.com/}{https://maqsam.com}.} for providing the compute to run some of our early experiments. We would like to thank Amgad Hasan and Maryam Abuflous for helping in the annotation process. Additionally, we want to thank Mohammed Almukhtar for sharing insights into the fine-tuning and inference of the \textsc{AceGPT-7B} model. 


\bibliography{anthology,custom}
\bibliographystyle{acl_natbib}

\appendix

\section{CIDAR Data Card} \label{appx:Data-Card}
We adopt the same template used by the NLLB team \cite{costa2022no}.
\subsection{Data Description}
\begin{itemize}
    \item Dataset Summary: \textit{ \textsc{Cidar} is a 10k culturally aligned dataset adopted from AlpaGusus.
    }
    \item Dataset Access:\textit{ You can access \textsc{Cidar} from Hugging Face Hub at }\href{https://hf.co/datasets/arbml/CIDAR}{hf.co/CIDAR}.
\end{itemize}
\subsection{Data Structure}
\textit{Dataset is uploaded as a single file in parquet format with 3 features: instruction, output, and index.}
\subsection{Data Creation}
\begin{itemize}
    \item Source Data: \textit{The dataset was created by selecting around 9,109 samples from \textsc{AlpaGasus} dataset and then translating it using ChatGPT. In addition, we appended that with around 891 instructions from the website Ask the Teacher. }
    \item Data Adoption: \textit{The 10,000 samples were reviewed by around 12 reviewers.}
\end{itemize}
\subsection{Considerations when using CIDAR}
\textsc{Cidar} \textit{is intended for research purposes only. The authors disclaim any responsibility for misuse and condemn any use contrary to Arabic culture or Islamic values. Even though subjected to human verification, there is no guarantee that responses are entirely aligned with Arabic culture and Islamic values. Users of the dataset are urged to exercise caution, employ critical thinking, and seek guidance from representative figures when necessary.}

\subsection{Additional Information}

\begin{itemize}
    \item  Dataset Curators: \textit{The authors of the paper.}  
    \item Licensing Information: \textit{The dataset is released under CC-BY-NC. The text and copyright (where applicable) remain with the original authors or
publishers, please adhere to the applicable licenses provided by the original authors.} 
    \item Citation Information: \textit{CIDAR Team et al, CIDAR: Culturally Relevant Instruction Dataset For Arabic, Arxiv, 2024.}
\end{itemize}

\begin{figure*}[!ht]
  \centering
\includegraphics[scale=0.51]{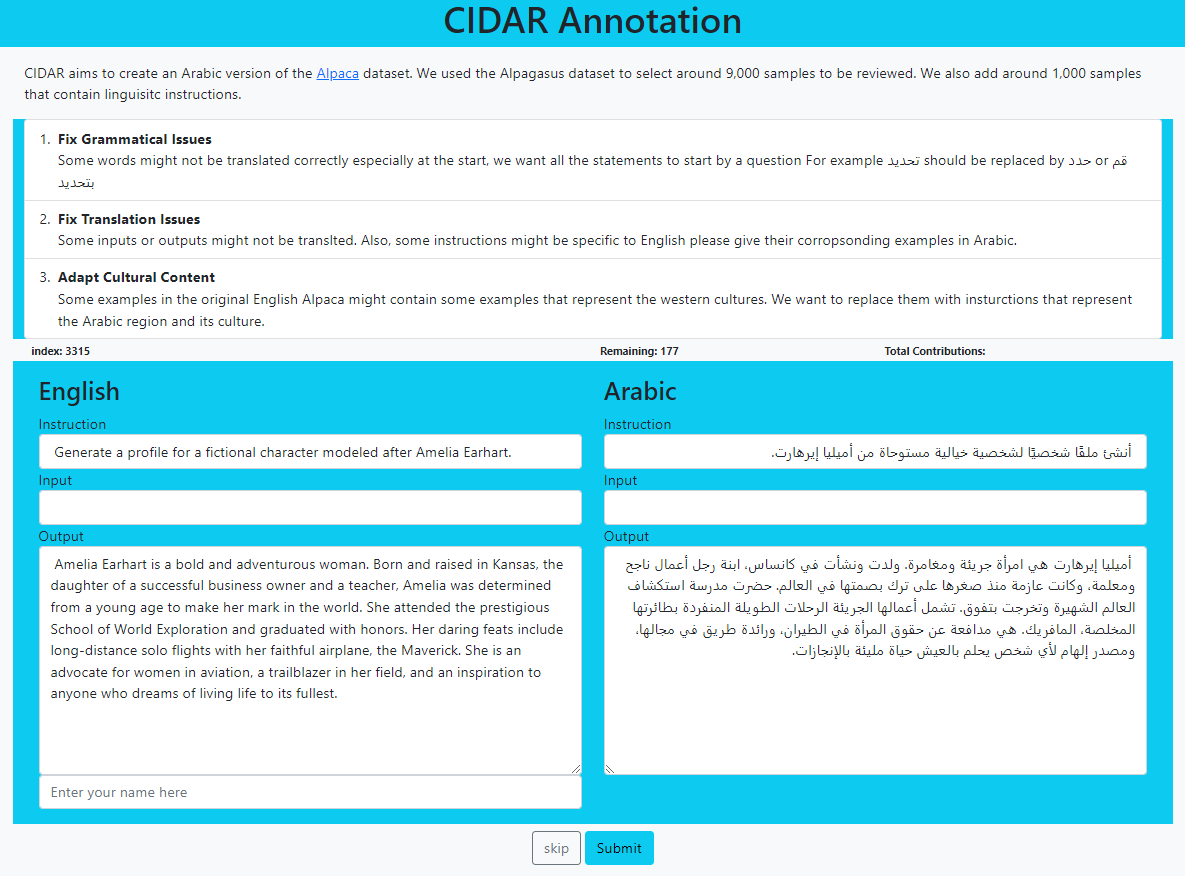}
 \caption{A screenshot of \textsc{Cidar} Annotation App, showing its features. The annotators can use it to fix grammatical issues, fix translation issues, and culturally localize a given instruction and output pair from any given dataset.}
    \label{fig:ann}
\end{figure*}
\section{Annotation App} \label{appx:Annotation-App}

The annotation app\footnote{Annotation App: \href{https://alpacaarabic-production.up.railway.app}{https://alpacaarabic.railway.app}.} contains two main parts for English and Arabic. Reviewers can make changes to \texttt{Instruction}, and \texttt{Output} to fix mistakes and align data with the Arabic culture. The original English instructions are shown to guide the reviewers for better re-annotation of the data. We have given the annotators 2 tasks (see Subsection \ref{localization}) that they should take into consideration during the annotation process. We require the annotators to write their names in the bottom left corner. The annotators can use \textit{Total Contributions} to keep track of their contributions to \textsc{Cidar} and \textit{Remaining} to keep track of the remaining samples to be re-annotated. We also allow the annotators to observe the reviewed submissions\footnote{Annotator Panel: \href{https://alpacaarabic-production.up.railway.app/explore}{https://alpacaarabic.railway.app/explore}.} and track the distribution of contributions. The website is designed using the Flask {framework}
\footnote{Flask Framework: \href{https://flask.palletsprojects.com/en/3.0.x/}{https://flask.palletsprojects.com}.}. The app regularly (every 1 hour) pushes the changes to the Hugging Face to save the progress. The website is deployed using Railway\footnote{Railway: \href{https://railway.app}{https://www.railway.app}.}.

\section{Instruction Datasets}
\label{appx:data}
In Table \ref{tab:rl}, we showcase the main instruction-tuning datasets that include Arabic subsets/versions from the literature. We highlight that to the best of our knowledge, all the datasets used to instruct-tuned Arabic LLMs are mostly machine-generated without human review or editing. 
\definecolor{Gray}{gray}{0.97}
\newcolumntype{A}{>{\raggedright\arraybackslash\columncolor{gray!10}}m{0.33\linewidth}}
\newcolumntype{B}{>{\centering\arraybackslash}m{0.15\linewidth}}
\newcolumntype{C}{>{\raggedright\arraybackslash}m{0.7\linewidth}}
\newcolumntype{D}{>{\centering\arraybackslash}m{0.06\linewidth}}
\newcolumntype{E}{>{\centering\arraybackslash}m{0.06\linewidth}}
\newcolumntype{F}{>{\centering\arraybackslash}m{0.33\linewidth}}
\renewcommand{\arraystretch}{1.5}

\begin{table*}[!ht]
\caption{Collection of Arabic instruction-tuning datasets discussed in the literature (Section \ref{sec:2}), highlighting their 
Arabic instructions count, dataset collection, type (multilingual or monolingual), and access status (open or closed).}
\label{tab:rl}
\begin{center}
\resizebox{\textwidth}{!}{\begin{tabular}{ | A | B | C | D | E| }

\hline

 \rowcolor{gray!10} \textbf{Dataset Name} & \textbf{Size (ar)} & \textbf{Dataset Collection} & \textbf{Type} & \textbf{Status} \\ \hline 

   xP3 \par \cite{muennighoff2022crosslingual}& 2,148,955 & Prompts applied to multiple datasets  & \parbox[t]{2mm}{\multirow{7}{*}{\rotatebox[origin=c]{90}{Multilingual}}} & \parbox[t]{2mm}{\multirow{8}{*}{\rotatebox[origin=c]{90}{Open}}}  \\ \cline{1-3}

   MSIFT \par \cite{Chen_MultilingualSIFT_Multilingual_Supervised_2023}& 114,231 &Translated using GPT4: Alpaca-GPT4, Evol-Instruct, ShareGPT  &  & \\ \cline{1-3}

  OASST1 \par \cite{köpf2023-openassistant}&  666& 
Conversational data was collected using a web app interface and obtained through crowd-sourcing. &  & \\ \cline{1-3}

 xP3x \par \cite{muennighoff2022crosslingual}& 18,246,158 & 
An extended large version of the xP3 dataset with multi-dialectal Arabic instructions, besides the Modern Standard Arabic instructions. &  & \\ \cline{1-3}

\textsc{SupNatInst} \par \cite{wang2022supernaturalinstructions}&  80,396& 
A large benchmark was collected through a large community effort on GitHub with the help of university students and NLP practitioners. &  &  \\ \cline{1-3}

  MITD \par \cite{upadhayay2023taco}& 81,451 & 
A composed multilingual instruction-tuning dataset from Alpaca-GPT4, Databricks’ Dolly, and Vicuna Benchmark in 132 languages, including Arabic, was translated using Google Cloud Translation.
 &  &   \\ \cline{1-3}

Bactrian-X \par \cite{li2023bactrianx}& 67,017 & Translated Alpaca using Google Translate then Feed to GPT3.5 Turbo. &  & \\ \cline{1-4}


  alpaca-arabic-instruct \par \cite{Alpaca_arabic_instruct}& 52,002 & Alpaca translated using Google Translate &  \parbox[t]{2mm}{\multirow{5}{*}{\rotatebox[origin=c]{90}{Monolingual}}} & \\ \cline{1-3} \cline{5-5}
\addlinespace[0.5ex]  
\cline{1-3} \cline{5-5}


 Jais Instructions \par \cite{sengupta2023jais}& 3,683,144& 
 xP3-Ar, Super-NaturalInstructions-Ar, Baize-Ar, Unnatural-Ar, Natural Questions-Ar, Bactrian-Ar, Alpaca-Ar, SafetyQA-Ar, NativeQA-Ar, Dolly-Ar, HC3-Ar, NER-Ar, Basic-Conv-Ar & & \parbox[t]{2mm}{\multirow{4}{*}{\rotatebox[origin=c]{90}{Closed}}} \\ \cline{1-3} 

   AceGPT Instructions \par \cite{huang2023acegpt} & 363,155 & 
Quora-Arabic, Alpaca-Arabic, Code-Alpaca-Arabic, Evol-Instruct-Arabic, ShareGPT.
   
&  & \\ \cline{1-3}

AlGhafa Instructions \par \cite{almazrouei-etal-2023-alghafa} &  1,459,000  & xP3-Ar, Bactrian-Ar, Alpaca-Ar, UltraChat-Ar &  &   \\ \cline{1-3}

 Noon Instructions \par \cite{naseej} & 110,000 &  Alpaca Instructions GPT4, Self-instruct records, Databricks, TruthfulQA, Grade School Math, Arabic-arithmetic-ChatGPT  &  &  \\ \cline{1-3} 

 Phoenix Instructions \par \cite{chen2023phoenix} & 8,000 
 &  A collection of translated Alpace instructions using GPT-4 to Arabic with a mixture of Arabic-generated responses for the GPT-4 translated instructions using GPT-3.5 Turbo. &  & \\
 
\hline
\end{tabular}}
\end{center}

\end{table*}

\section{Used Hyper-parameters}
\label{appx:params}
This section provides detailed specifications of the hyper-parameters used in the inference and fine-tuning of the \textsc{AceGPT}-7B model.

\begin{table*}[!ht]
\begin{center}
\caption{List of the fine-tuning 
parameters for the models fine-tuned on \textsc{Cidar} and the translated \textsc{AlpaGasus}.}
\resizebox{\textwidth}{!}{\begin{tabular}{ | A | F | A | F | }
\hline
\textbf{Parameter} & \textbf{Value} & \textbf{Parameter} & \textbf{Value} \\ \hline \hline
lora\_r & 16 & lora\_alpha & 16 \\
lora\_dropout & 0.1 & bnb\_4bit\_compute\_dtype & "bfloat16" \\
bnb\_4bit\_quant\_type & "nf4" & bf16 & True \\
num\_train\_epochs & 3 & per\_device\_train\_batch\_size & 2 \\
per\_device\_eval\_batch\_size & 2 & gradient\_accumulation\_steps & 1 \\
gradient\_checkpointing & True & max\_grad\_norm & 0.3 \\
learning\_rate & 2e-4 & weight\_decay & 0.001 \\
optim & "paged\_adamw\_32bit" & warmup\_ratio & 0.03 \\
group\_by\_length & True & & \\ \hline
\end{tabular}}

\label{table:finetune_param}
\end{center}
\end{table*}

Table \ref{table:finetune_param} details the fine-tuning hyper-parameters employed to optimize the models' performance. It includes adjustments to learning rates, batch sizes, and regularization, alongside LoRA adaptations and precision formats. Specifically, we loaded the models in 4-bit precision and used for LoRa a low rank ($r$) of 16 and a scaling factor (alpha $\alpha$) of 16.

In the inference setup, we used the \texttt{text-generation} pipeline from HuggingFace\footnote{Pipelines: \href{https://hf.co/docs/transformers/main\_classes/pipelines}{https://hf.co/docs/transformers/main\_classes/pipelines}.} with the following hyper-parameters: \texttt{max\_length=512} to constrain output length, \texttt{temperature=0.2} for lower randomness favoring higher probability tokens, \texttt{top\_p=1.0} and \texttt{top\_k=0} allowing full probability distribution without restricting to top tokens, \texttt{repetition\_penalty=1.2} to reduce repetition, and \texttt{do\_sample=True} to enable stochastic sampling. These settings were chosen carefully to balance coherence and context relevance, aligning with our objectives for high-quality and diverse linguistic output.


\section{Example Outputs}
\label{appx:outputs}

In Table \ref{fig:outputs}, 
we give some example outputs for a few given Arabic instructions generated by the three evaluated models (\textsc{AceGPT}\textbackslash \textsc{Cidar}, \textsc{AceGPT}\textbackslash \textsc{AlpaGasus}, and \textsc{AceGPT}\textbackslash \textsc{Chat}) used in this study, like `How did our language originate? \<كيف نشأت لغتنا؟> '. To prevent any bias, we use the same inference parameters for all the models. Furthermore, we do not generate multiple outputs or cherry-pick specific outputs for the same instruction. We provide the outputs considering various topics, like clothes, fonts, food and drinks, language, grammar, and traditions. It is clear that the examples provided show that \textsc{AceGPT}\textbackslash \textsc{Cidar} can better capture the culture compared to the \textsc{AceGPT}\textbackslash \textsc{Chat}, which was fine-tuned on hundreds of thousands of instructions, whereas  \textsc{AceGPT}\textbackslash \textsc{AlpaGasus} produces the worst results in terms of cultural relevance and grammar.

\newgeometry{left = 2.5cm, top = 1cm}

\includepdf[pages={1-4}]{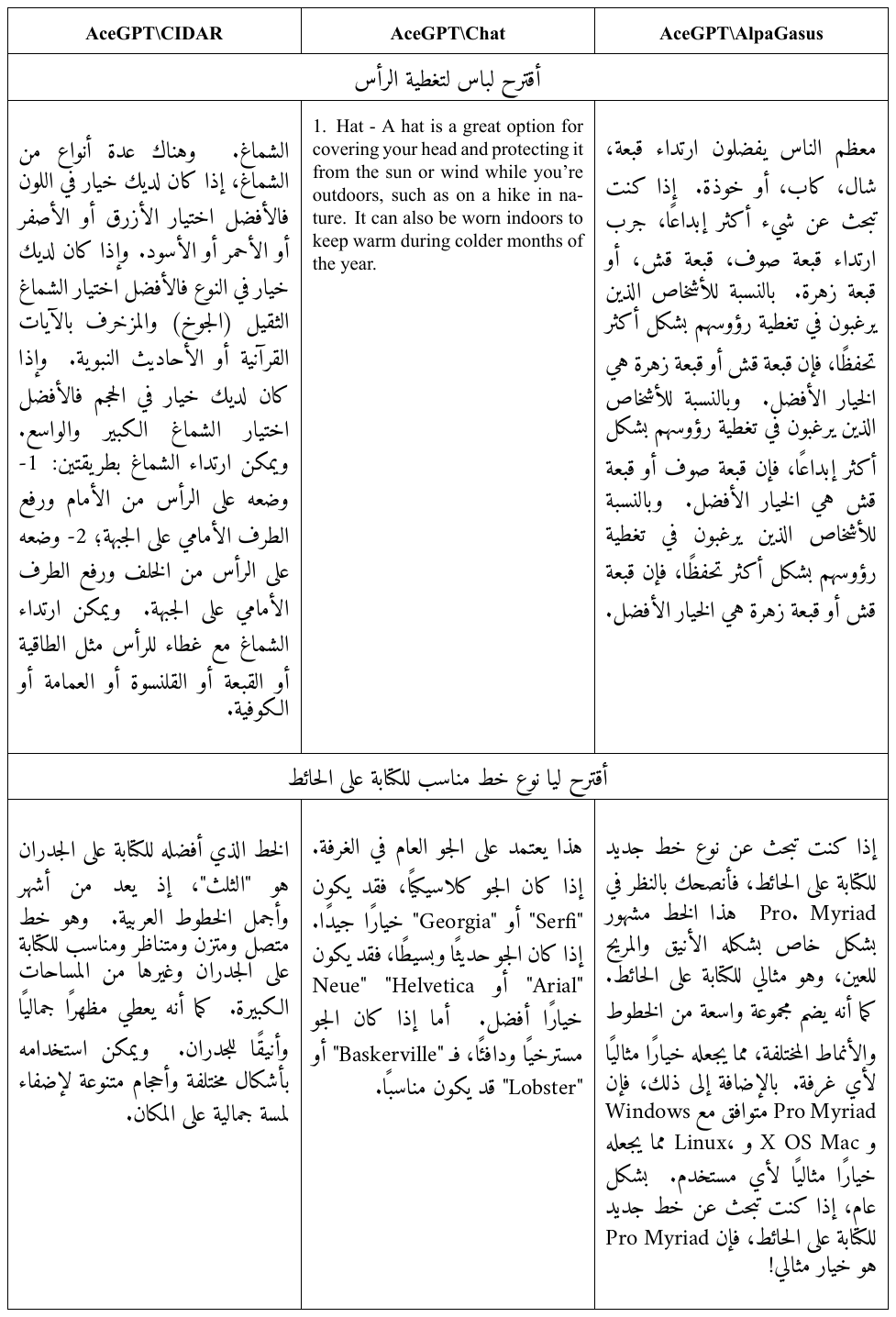}

\begin{table}[!htp]
\begin{center}
\begin{tabular}{c}
\includegraphics[page=5,width=1.15\textwidth,trim={3.0cm 15.5cm 0 0},clip]{imgs/examples.pdf}
\end{tabular}
\caption{Example outputs for a few given Arabic instructions generated by the three evaluated models (\textsc{AceGPT}\textbackslash \textsc{Cidar},
\textsc{AceGPT}\textbackslash \textsc{AlpaGasus}, and
\textsc{AceGPT}\textbackslash \textsc{Chat}) used in this study. We note that some sentences have been truncated for better readability.}
\label{fig:outputs}
\end{center}
\end{table}

\end{document}